# Traffic event description based on Twitter data using Unsupervised Learning Methods for Indian road conditions

K. Yasaswi Sri Chandra Gandhi[*1], Indrajit Ghosh[2]

[1, 2] Civil Engineering Department, Indian Institute of Technology Roorkee, India
(E-mail: kgandhi@ce.iitr.ac.in, indrafce@iitr.ac.in)

***Short abstract***: Non-recurrent and unpredictable traffic events directly influence road traffic conditions. There is a need for dynamic monitoring and prediction of these unpredictable events to improve road network management. The problem with the existing traditional methods (flow or speed studies) is that the coverage of many Indian roads is very sparse and reproducible methods to identify and describe the events are not available. Addition of some other form of data is essential to help with this problem. This could be real-time speed monitoring data like Google Maps, Waze, etc. or social data like Twitter, Facebook, etc. In this paper, an unsupervised learning model is used to perform effective tweet classification for enhancing Indian traffic data. The model uses word-embeddings to calculate semantic similarity and achieves a test score of 94.7%.

## 1. INTRODUCTION

Traffic events can be divided into those with recurrent predictable causes and non-recurrent unpredictable causes. These non-recurrent events, consisting of traffic incidents, unplanned roadworks, weather, and special events, which are critical but difficult to detect, categorize and describe. These traffic events are of interest to multiple groups. First, teams that are interested in traffic event detection. Second, people that have to apply traffic event management. Third, organizations that use an offline historical analysis of traffic events and carry research to understand the causes behind and statistics on the circumstances. The applications can be used to study the past, manage the present conditions effectively and predict future events with better accuracy than we are used to seeing currently. These organizations have multiple tools at their disposal to achieve their tasks, such as roadside detection sensors and cameras. Also, they have the availability of information provided by traffic inspectors and emergency services. [1][2][3] However, the current tooling problem is that their capabilities to detect, categorize, and describe traffic events are limited and flawed.

The problems involved are two-fold; most existing algorithms assume that traffic events cause an immediate change in traffic flow and description. This might not necessarily be the case always. The reason it happens is that the data on which these algorithms depend is provided by traffic sensors, which are limited in amount and cannot cover every point on the road, in other words, the data is very sparse. [4] The other issue is that most of the existing methods are heavily dependent on the type of road. The same events on arterial roads will not have a similar effect on freeways. [5]

*Corresponding Author*

## 2. OBJECTIVES

To tackle this problem, Twitter data [6] is used to enhance the already existing traditional data and improve the detection, categorization and description of traffic events. Irrespective of the data source, the data has been underlined as obtained from willing individuals who have knowingly and voluntarily put up traffic event data, which also has attached to it some sort of location properties. As dynamic social sensors, people might immediately report traffic conditions (e.g., slow-moving traffic) and traffic incidents (e.g., accidents) by posting tweets while they are driving or observing the occurrence of a traffic-related event. Accordingly, Twitter data can be considered a complementary traffic data source and traditional sensors such as CCTV cameras and inductive loop detectors.

## 3. MATERIALS AND METHODS

This section will give a concise overview of the data collection methods and the unsupervised learning algorithm used to tackle this problem. The solution that worked most efficiently and practically was the word-embedding approach. Word-embedding techniques map millions of words and phrases into numerical vector space in such a way that semantic-and-syntactic similar names are closer to each other. Initially, a small traffic keyword set is developed by extracting the most frequently tweeted words from a small bucket of the tweets that we have manually picked. Word-embedding tools are used to measure a heuristic in vector space and check whether a sample tweet exceeds that heuristic. All tweets that exceed that number are classified under traffic-related tweets. The only parameter that training happens against is the threshold heuristic. This is the power of the word-embeddings method.

### 3.1. Data Collection

The main goal while collecting social media data is that a small keyword set needs to be created that maximizes the number of tweets with respect to the overall acquired tweets and also the tweets in the entire universal set as far as non-recurrent traffic events are concerned. Mathematically, this means precision and recall need to be maximized. The tweets were divided into three categories, 1) Unrelated to traffic, 2) Traffic-related high priority (containing mostly traffic-related keywords) and 3) Traffic-related low priority (everyday events like traffic jams, weather events.)

The collected traffic-related tweets were tokenized, and the token occurrence frequency was computed for each distinct token. Finally, tokens with the highest occurrence frequency were selected to build the traffic-related dictionary. Examples of traffic words in the dictionary were "caves, heavy, rain, one-way, non-fatal, accident, congestion, delays, vehicle, incident, crash, and road". [7]

### 3.2 Unsupervised Learning Methods

The problem at hand is a binary classification problem, as tweets need to be classified into one of the two categories where one is traffic-related, and the other is not. The traffic-related tweets are to be further divided into two types, ones who report traffic events like crashes etc. and ones who report traffic conditions like slow-moving traffic etc. Identification of three major tasks to accomplish was done and the same are listed here, 1) Identify traffic keywords, 2) Obtain vector representations for words and 3) Classification.

A similarity heuristic in vector space was used for differentiating between traffic-related tweets and the other regular tweets. Google's word2vec [8] model was employed for this task, as millions of words are mapped to vectors of real numbers. In this representation, words with similar semantic meaning are closer to each other than the rest [9].

## 4. RESULTS

The results show that certain grammatical features were extracted from the tweets which were helping the models to classify the inference data better.

Example tweet: Optional(Too much, because of, how much more) + Optional (jam, road work, white topping) + Roadwork token + Optional (CM convoy, Chinnaswamy stadium rush, RCB victory celebrations) + Optional(Swift, Bishop Cottons bus, White SUV) + Optional (Autowala, bullock cart, pothole)

On training the model and computing the threshold on the Google word2vec model, testing was done on the Twitter dataset, which resulted in an accuracy of 94.7%. The same test accuracy with less than 1% variation was achieved by our methods when trained on separated training sets with two types of word2vec models.

## 5. CONCLUSION

The power of word-embedding models and Google's word2vec model were used and trained to develop a simple but useful framework to classify tweets. The main features of this model are:

1. Model is primarily based on unsupervised learning mechanisms
2. The model only trains one parameter which is the threshold
3. Only a small dictionary of keywords is necessary, and the model itself proves to be independent of the sample size when the sample quality is good enough.
4. The metrics show promise with a few being high test accuracy, precision, recall and F1 score.
5. **These results can be used to further improve existing statistical models for Indian roads. [10]**